\pdfoutput=1

\documentclass[11pt]{article}

\usepackage[final]{acl}

\usepackage{times}
\usepackage{latexsym}

\usepackage[T1]{fontenc}

\usepackage[utf8]{inputenc}

\usepackage{microtype}

\usepackage{inconsolata}

\usepackage{graphicx}
\usepackage{fontawesome}

\usepackage{fancyvrb}
\DefineVerbatimEnvironment{Code}{BVerbatim}{baseline=t}

\usepackage{multirow, makecell}
\newcommand{\mail}[1]{{\fontsize{11pt}{11pt}\selectfont\href{mailto:#1@udc.es}{\texttt{#1@udc.es}}}}
\title{LyS at SemEval 2025 Task 8: Zero-Shot Code Generation for Tabular QA}

\author{
    Adrián Gude \\ \mail{adrian.lopez.gude}  \And  \textbf{Roi Santos-Ríos}\\  \mail{roi.santos.rios} \\ \AND \textbf{Francisco Prado-Valiño} \\ \mail{francisco.prado.valino} \\ \And \textbf{Ana Ezquerro} \\ \mail{ana.ezquerro} \\[0.5em] Universidade da Coruña, CITIC \\ Departamento de Ciencias de la Computación y Tecnologías de la Información \\ Campus de Elviña s/n, 15071, A Coruña, Spain \\ \And Jesús Vilares\\  \mail{jesus.vilares} 
}

\usepackage{fixltx2e}

\begin{document}
\maketitle
\begin{abstract}
This paper describes our participation in SemEval 2025 Task 8, focused on Tabular Question Answering. We developed a zero-shot pipeline that leverages an Large Language Model to generate functional code capable of extracting the relevant information from tabular data based on an input question. Our approach consists of a modular pipeline where the main code generator module is supported by additional components that identify the most relevant columns and analyze their data types to improve extraction accuracy. In the event that the generated code fails,  an iterative refinement process is triggered, incorporating the error feedback into a new generation prompt to enhance robustness. Our results show that zero-shot code generation is a valid approach for Tabular QA, achieving rank 33 of 53 in the test phase despite the lack of task-specific fine-tuning. 

\end{abstract}

\section{Introduction}
    Tabular Question Answering (Tabular QA) has huge potential in real-world applications such as financial analysis, business intelligence, and scientific data exploration, where structured databases serve as the primary source of information. Unlike traditional text-based Question Answering (QA), which primarily deals with unstructured data, Tabular QA requires extracting information from structured tables to be able to answer the input questions, thus involving reasoning about diverse table schemas, column relationships, and heterogeneous data types.

    Complex supervised systems have been proposed to deal with the structured nature of Tabular QA, either leveraging structured prediction with language representations~\cite{herzig-etal-2020-tapas, yin-etal-2020-tabert} or by formulating the task as a sequence-to-sequence problem~\cite{zhong2017seq2sqlgeneratingstructuredqueries, yu-etal-2018-typesql, pal-etal-2023-multitabqa}. However, with the rise of instruction-based Large Language Models (LLM)~\cite{brown-etal-2020-language}, recent approaches have shifted away from reliance on large annotated datasets, instead reframing the task as a zero-shot generation problem~\cite{cao-etal-2023-api}.
    
    In this work, we further explore instruction-based LLMs to dynamically generate code functions capable of retrieving relevant data from tables based on the input question in a zero-shot manner. To enhance accuracy and reliability, we developed a modular three-staged pipeline that includes: (i)~a column selection mechanism to determine the most relevant columns and their data-type, (ii)~a code generation module responsible for producing executable code and (iii)~an iterative error handling module that, in case the initial code execution fails, tries to fix the generated code accordingly.
    
    Our group tested this approach within the SemEval 2025 Task 8 event~\cite{osesgrijalba-etal-2025-semeval-2025}, which provided a diverse dataset featuring real-world tabular data.\footnote{Our implementation is fully available at {\url{https://github.com/adrian-gude/Tabular_QA}} (Feb. 2025).} The competition required models to produce answers in multiple formats, including boolean, categorical, numerical, and list-based outputs. Our model was designed to generalize across different table structures, making it adaptable to various datasets beyond the shared task, ensuring robustness and broad applicability. Although our approach demonstrated strong performance in code generation and execution, subsequent analysis revealed that the model struggles with columns containing complex data types (lists, dictionaries, etc.) and ambiguous queries, particularly for list-based responses.

\section{Background}

    Question Answering (QA) has been gaining significant attention in recent years, driven by the need for models capable of reasoning over structured data. Early tasks in QA mainly focused on retrieving information from unstructured text sources~\cite{rajpurkar2016squad100000questionsmachine, yang-etal-2018-hotpotqa}, but the increasing availability of structured datasets has led to new challenges in understanding and querying tabular data.
    Unlike classic text-based QA, where answers are retrieved from free-form text, Tabular QA requires a higher level of interpretation and robustness to map questions to relevant columns and rows, handle missing values, and compute statistics when necessary.
            
    In parallel, several datasets have been introduced to benchmark Tabular QA models, including WikiTableQuestions~\cite{pasupat-liang-2015-compositional}, SQA~\cite{iyyer-etal-2017-search}, and the more recent DataBench dataset~\cite{oses-etal-2024-databench}, which provides real-world tabular data for evaluating models in different scenarios.

    \paragraph{Structured Tabular QA}
    Most state-of-the-art approaches for Tabular QA leverage a pretrained language model ---equipped with an specialized encoding module to represent tabular information--- tailored for structured prediction. For example, \textsc{TaPas}~\cite{herzig-etal-2020-tapas} feeds both the input question and the flattened table into BERT~\cite{devlin-etal-2019-bert} as a single sequence, and finetunes the architecture to select relevant columns and predict an aggregation function.

    Similarly, \textsc{TaCube}~\cite{zhou-etal-2022-tacube} combines a cube constructor with BART~\cite{lewis-etal-2020-bart} to predict the real answers based on the input question and the results of the cube operations.

    \paragraph{Generative Tabular QA}
        To address the rigidity of structured approaches, recent works have explored generative models for program synthesis, where an LLM is finetuned to generate executable programs or instructions (in the form of SQL queries, for example) to be applied against tabular sources.  \citet{zhong2017seq2sqlgeneratingstructuredqueries} proposed \textsc{Seq2SQL}, a sequence-to-sequence model to translate natural language into SQL syntax, incorporating query-space pruning to significantly simplify and enhance the generative task. Later, \citet{yin-etal-2020-tabert} joined both concepts by optimizing tabular embeddings that fit both generative and structured purposes.

    \paragraph{Zero-Shot Code Generation}
        More recently, advancements in code generation have enabled a paradigm shift in Tabular QA, driven by powerful multipurpose LLMs with strong coding capabilities, such as Qwen~\cite{bai2023qwentechnicalreport} and Mistral's Codestral~\cite{jiang-etal-2023-mistral7b}.
         These models facilitate a zero-shot approach to program synthesis, eliminating the need for predefined templates or 
         large annotated datasets. Instead, zero-shot generation allows the system to dynamically adapt to different schemes without explicit prior knowledge of the table structure~\cite{cao-etal-2023-api}, thus providing flexibility and scalability. 
        
        Despite its potential, zero-shot code generation models still face big challenges, particularly in error handling, runtime execution failures, and schema variability. Building on this approach,  our work extends an instruction-based model with error awareness, enabling it to detect and recover from execution failures in an iterative error-recovery mechanism, where the model dynamically analyzes execution failures and regenerates code based on error feedback.

\section{System Overview}
Our approach for the SemEval 2025 Task 8 iterates upon the code generation approaches for Tabular QA, where the core component is a pretrained LLM responsible of generating executable code to extract the answer from the tables. To build upon prior works~\cite{herzig-etal-2020-tapas}, we incorporated a module that helps selecting the columns relevant to the question, while also identifying the data types of their content. Moreover, we incorporate an error-fixing module that attempts to catch runtime errors and integrates them as part of a new prompt, guiding the LLM to refine its code generation.

\begin{figure*}[hbtp!]
        \includegraphics[width=\textwidth]{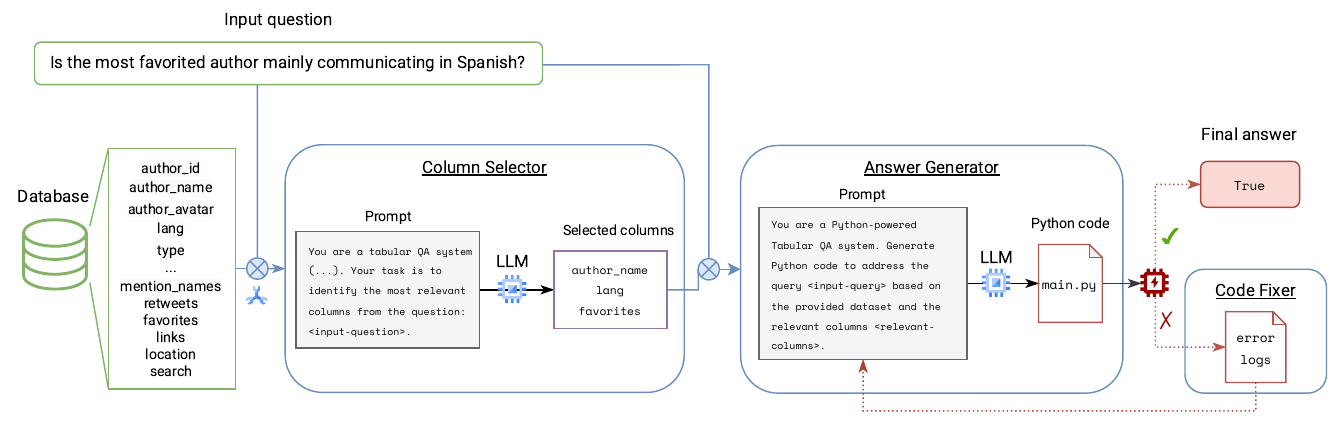}
        \caption{Architecture of our system. Different symbols are used to represent different elements of our pipeline: $\otimes$ merges information in a prompting-like form, \includegraphics[height=0.3cm]{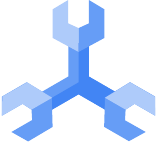} represents a preprocessing step, \includegraphics[height=0.3cm]{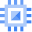} indicates LLM inference (with optional post-processing steps), and \includegraphics[height=0.3cm]{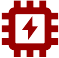} runs Python code and catches error logs. Solid lines are used to indicate fixed pipeline steps while dotted lines indicate optional steps that are executed depending on partial results of the system. Green boxes represent elements provided in the task.}
        \label{fig:experiments}
    \end{figure*}

    Figure~\ref{fig:experiments} shows an schematic view of the architecture of our system. We have designed a modular pipeline that features three main components, which we describe below: (i)~a column selector, (ii)~an answer generator and (iii)~a code fixer. 
  
    \paragraph{Column Selector} Instead of relying on manually crafted heuristics or embedding similarity measures, the first component of our system leverages an instruction-based LLM tasked to identify the most relevant columns of a tabular source from an input question in natural language form. 
    Our template provides the list of column names and instructs the model to return only those that are essential for answering the query.\footnote{All our prompts are available in the code publicly available at GitHub.} 

    \paragraph{Answer Generator} Once the relevant columns are identified, the second component of our pipeline is instructed to generate executable code that retrieves the answers from the tabular source using both the input query and the relevant columns extracted in the previous step. As part of our prompt, we guided the LLM to generate Python programming code and postprocessed the output to ensure that only Python lines were passed throught the next module. Python language was chosen since it is widely used in data analysis and has extensive support for tabular data processing through libraries such as Pandas. 

    \paragraph{Code Fixer}  The final component of our pipeline captures execution errors that might occur due to incorrect syntax,  schema mismatches, or runtime exceptions. This module captures the error messages and re-generates a corrected function by feeding the error context back into the LLM. To achieve this, we used a structured prompt that includes the code that causes an error with the corresponding error description.

    \paragraph{Preprocessing}  Since our system strongly relies on a well-formatted prompt, we manually designed a preprocessing step to ensure a consistent format to feed our system.  We standardized column names for simplified versions (removing emoji and all non-alphanumerical characters except punctuation symbols) to prevent possible errors in the Answer Generator caused by mismatches between the table structure and the generated code. We identified enum-like column types, such as the case of categorical attributes with a finite amount of strings as a value (e.g. a ``\textit{Survey}'' column that only contains ``\textit{Yes}'', ``\textit{No}'' or ``\textit{Maybe}''), and inferred a common scheme so to ensure consistency across different attributes, thus reducing errors related to unexpected variations in categorical values.

\section{Experimental Setup}

        Our system relies on open-source LLMs for zero-shot code generation. This way, no explicit training nor finetuning was conducted. Instead, we used the available training phase datasets to validate different LLMs and select the best performing one for the final test phase.
        
    \paragraph{Dataset} The dataset provided for the task is divided into three sets: \textit{training}, \textit{development} (aka \textit{dev}), and \textit{test}.     
    In our case, since we had opted for a zero-shot approach, the training set remained unused during the development phase, using only the dev set for our experiments.     During this stage we tried different LLMs to compare their ability to generate the adequate Python code to answer the input questions. To do that, we analyzed the accuracy obtained with respect to the ground truth of the validation set, together with manual checks to assess the quality of the generated code.

        \paragraph{Evaluation} 
        The official evaluation consists of two subtasks, where \textit{Subtask~1} uses all available data sources to answer the input question, while \textit{Subtask~2} operates on a limited database, sampling a maximum of 20 rows per table to perform queries.

        \paragraph{System Setup} We conducted experiments with different open-source LLMs adjusted to our hardware limitations, specifically pretrained for instruction-based code generation: Qwen-2.5-Coder~\cite{bai2023qwentechnicalreport} (with~\href{https://huggingface.co/Qwen/Qwen2.5-Coder-7B-Instruct}{7B} and~\href{https://huggingface.co/Qwen/Qwen2.5-Coder-32B-Instruct}{32B} versions), \href{https://huggingface.co/mistralai/Mistral-7B-Instruct-v0.3}{Mistral-7B} and~\href{https://huggingface.co/mistralai/Codestral-22B-v0.1}{Codestral-22B} ---the later two from Mistral~\cite{jiang-etal-2023-mistral7b}.

        To run the generated code  we relied on Python~3.10.12 with Pandas~2.2.3 as a requirement. Due to VRAM constraints, all models were executed with 4-bit quantization, using a greedy generation strategy with a temperature of~0.7.

\section{Analysis of Results}
    In this section, we present the evaluation of our system on the task. We first report performance during the development phase (\$\ref{subsec:dev}), where we experimented with different models on the validation dataset, followed by the final test phase (\$\ref{subsec:test}), where our system was evaluated on the test dataset through CodaBench submissions.\footnote{\scriptsize\url{https://www.codabench.org/competitions/3360/}.}
    
    \subsection{Development Phase}
    \label{subsec:dev}

        As explained before, during the development phase we focused on selecting the best performing LLM just using the dev set; that is, dismissing the training set. 
        At this first stage, our pipeline was conformed by only the Answer Generator module.

        \begin{table*}[htbp!]
          \centering\small
          \setlength{\tabcolsep}{5pt}
          \begin{tabular}{|l|l|rrrrr|r|c|}
            \cline{3-9}
            \multicolumn{2}{c|}{}& \textbf{boolean} & \textbf{category} & \textbf{number} & \textbf{list[category]} & \textbf{list[number]} & \makecell[c]{$\mu$} & $\beta$\\
            \hline 
            \parbox[t]{2mm}{\multirow{4}{*}{\rotatebox[origin=c]{90}{\textbf{S1}}}} 
                & Qwen-2.5-Coder\textsuperscript{7B} & 67.19 & 68.75 & 75.00 & 3.12 & 3.12 & 43.44 & \multirow{4}{*}{27.00}\\
                & Mistral\textsuperscript{7B}& 51.56 & 59.37 & 73.44 & 35.94 & 34.37  & 50.94 &\\
                & Codestral\textsuperscript{22B} & 73.44 & \textbf{82.81} & \textbf{82.81} & 48.44 & 48.44   & 67.19 & \\
                & Qwen-2.5-Coder\textsuperscript{32B} & \textbf{81.25} & 78.12 & 75.00 & \textbf{65.62} & \textbf{70.31}  & \textbf{74.06} & \\
            \hline
            \parbox[t]{2mm}{\multirow{4}{*}{\rotatebox[origin=c]{90}{\textbf{S2}}}} 
                & Qwen-2.5-Coder\textsuperscript{7B} & 81.25 & 84.37 & 85.93 & 6.25 & 1.56 & 51.87 & \multirow{4}{*}{26.00}\\
                & Mistral\textsuperscript{7B}  & 46.87 & 56.25 & 65.62 & 32.81 & 25.00  & 45.31&\\
                & Codestral\textsuperscript{22B} & 71.87 & \textbf{89.06} & 84.37 & 53.12 & 60.94 & 71.87  &\\
                & Qwen-2.5-Coder\textsuperscript{32B}  & \textbf{84.37} & \textbf{89.06} & \textbf{85.94} & \textbf{75.00} & \textbf{75.00} & \textbf{81.87}&\\
            \hline 
          \end{tabular}
          \caption{Performance of different LLMs on the validation set for Subtasks~1 and~2 (\textit{S1} and~\textit{S2}, respectively), where the pipeline only contains the Answer Generator module. Columns~$\mu$ and~$\beta$ indicate the average and baseline performance, respectively. The best performance is highlighted in bold.}
          \label{tab:train_results}
        \end{table*}
    
        The results obtained for this original setup, presented in Table~\ref{tab:train_results}, show that larger models such as Qwen-2.5-Coder\textsuperscript{32B} significantly outperform smaller models, with accuracy gains of over 20 points compared to Qwen2.5-Coder\textsuperscript{7B}. Regardless of the selected model, our zero-shot approach consistently outperforms the baseline system~\cite{osesgrijalba-etal-2025-semeval-2025} in both subtasks. Evaluation metrics indicate higher scores for Subtask~2 than for Subtask~1, likely due to the smaller input size, which reduces the amount of information introduced in the prompt and minimizes potential ambiguities when execution the generated code. We also notice a performance drop when breaking down the accuracy by the datatype, where even the best LLM struggles when generating answers for categorical list-like  attributes.

    \paragraph{Ablation Study}
        We relied on the results displayed in Table~\ref{tab:train_results} to select the best performing LLM, which served as the foundation for integrating the additional modules that could further enhance performance (see Figure~\ref{fig:experiments}). Table~\ref{tab:tabal} shows the results when varying the components of the pipeline while maintaining Qwen-2.5-Coder\textsuperscript{32B} as backbone.  
        The AG (Answer Generator only) setup corresponds to the result displayed in Table \ref{tab:train_results}, from which the extra components of our pipeline where compared to see if there was an actual improvement when introducing error-awareness and column pre-selection. The AG+CS (AG with Column Selector) setup shows a clear improvement of 3~and 2~points in each subtask with respect to the AG-only model, outlining the importance of first asking the LLM to filter the relevance of the input attributes. Lastly, when integrating the Code Fixer (CF) with an enhanced column selection (ECS) to feed richer information about feature variations to the prompt, our final system setup (AG+ECS+CF) maintains almost the same performance over Subtask~2 but improves 7~points in Subtask~1, proving that integrating error feedback to the model assists the LLM for better querying larger databases. Specifically, the largest performance boost is obtained in categorical list-like attributes, where the accuracy increases 10~points with respect to the AG+CS model.

        \begin{table*}[htbp!]
          \centering\small
          \setlength{\tabcolsep}{5pt}
          \begin{tabular}{|l|l|rrrrr|r|}
            \cline{3-8}
            \multicolumn{2}{c|}{}& \textbf{boolean} & \textbf{category} & \textbf{number} & \textbf{list[category]} & \textbf{list[number]} & \makecell[c]{$\mu$} \\
            \hline 
            \parbox[t]{2mm}{\multirow{3}{*}{\rotatebox[origin=c]{90}{\textbf{S1}}}} 
                & AG & 81.25 & 78.12 & 75.00 & 65.62 & 70.31 & 74.06 \\
                & AG+CS & 82.81 & 78.12 & 78.12 & 68.75 & 79.69 & 77.50 \\
                & AG+ECS+CF & \textbf{89.06} & \textbf{85.94} & \textbf{85.94} & \textbf{78.12} & \textbf{85.94} & \textbf{85.00} \\
            \hline
            \parbox[t]{2mm}{\multirow{3}{*}{\rotatebox[origin=c]{90}{\textbf{S2}}}} 
                & AG & 84.37 & \textbf{89.06} & 85.94 & 75.00 & 75.00 & 81.87 \\
                & AG+CS & 84.37 & \textbf{89.06} & \textbf{90.62} & 73.44 & \textbf{79.69} & 83.44 \\
                & AG+ECS+CF & \textbf{89.06} & \textbf{89.06} & \textbf{90.62} & \textbf{76.56} & 78.12 & \textbf{84.69} \\
            \hline 
          \end{tabular}
          \caption{Performance on the validation set for Subtasks~1 and~2 (\textit{S1} and~\textit{S2}, respectively) when integrating different components of the pipeline with Qwen-2.5-Coder\textsuperscript{32B} as backbone. The best performance is highlighted in bold.}
          \label{tab:tabal}
        \end{table*}

    \subsection{Final Test Phase}\label{subsec:test}
        The best performing configuration (AG+ECS+CF) was selected to participate in the competition. 
        Our zero-shot approach reached 65 points of accuracy in Subtask~1 and 68 points in Subtask~2. So, we ranked in the 32th (Subtask~1) and 31th (Subtask~2) positions out of 49 participants in the \textit{General} category, and 23th (Subtask~1) and 21th (Subtask~2) positions out of 35 participants in the \textit{Open models} category.
        
        Our results during the development phase (84 and 85 points for Subtasks~1 and~2, respectively) suffered a significant drop of 20 points (approx.) in accuracy with respect to the validation results, likely due to the greater complexity of datatypes presented in the test tables. For instance, the test set presents 
        multiple columns with
        lists that are not enclosed by square brackets, or that have variable separators for their elements (commas or semicolons); 
        and dictionaries with a variable amount of keys.\footnote{For example, a cell of the form: \texttt{Education;Social Protection;Agriculture, Fishing and Forestry} or \texttt{\{'service': 5.0, 'cleanliness': 5.0, 'overall': 5.0, 'value': 4.0, 'location': 5.0\}}.}
        Tables~\ref{tab:train_results} and~\ref{tab:tabal} show a clear difference in terms of accuracy when considering more complex datatypes: boolean accuracy reaches more than 80 points, while list-like types do not surpass 75 points. This might indicate that the LLM 
        is not able to infer these complex schemes on the test set, producing errors that are propagated from the Column Selector module to the Answer Generator.

\section{Conclusions and Future Work}
    In this work we propose a zero-shot approach for Tabular QA that demonstrated a strong performance for the SemEval 2025 Task 8, ranking among the best systems in the development phase, although suffering from a performance drop in the test phase. Still, our system shows that an instruction-based approach allows to dynamically adapt to different dataset schemes without requiring additional training or finetuning,  surpassing the baseline model even with limited hardware resources available.
    
    Future work will focus on further refining prompt templates, improving schema adaptation, optimizing execution efficiency or incorporating a voting system with different LLMs. Improving the detection of these complex datatypes is also critical, as they allow the model to answer questions on less structured tables ---which constitute the majority of online data---, ultimately making the system more generalizable.

\section*{Hardware Setup}
Our hardware resources are somewhat limited by today's standards. We had shared access to an Intel Core i9-10920X at 3.50~GHz with 258~GiB RAM and two integrated NVIDIA RTX~3090, so we opted to perform zero-shot instead of finetuning the LLMs. 

\section*{Acknowledgments}

We acknowledge grants SCANNER-UDC (PID2020-113230RB-C21) funded by MICIU/AEI/10.13039/501100011033; GAP (PID2022-139308OA-I00) funded by MICIU/AEI/10.13039/501100011033/ and ERDF, EU; LATCHING (PID2023-147129OB-C21) funded by MICIU/AEI/10.13039/501100011033 and ERDF, EU; CIDMEFEO funded by the Spanish National Statistics Institute (INE); as well as funding by Xunta de Galicia (ED431C 2024/02), and Centro de Investigación de Galicia ``CITIC'', funded by the \textit{Xunta de Galicia} through the collaboration agreement between the \textit{Consellería de Cultura, Educación, Formación Profesional e Universidades} and the Galician universities for the reinforcement of the research centres of the Galician University System (CIGUS).




\end{document}